\documentclass[letterpaper, 10 pt, conference]{ieeeconf}  
\pdfoutput=1

\IEEEoverridecommandlockouts                              

\usepackage{amsmath}
\usepackage{amsfonts}
\usepackage{amssymb}
\usepackage{mathtools}

\usepackage{graphicx}
\usepackage{epstopdf}
\usepackage{subcaption}
\DeclareCaptionLabelSeparator{periodspace}{.\quad}
\usepackage{multirow}
\newcommand\Tstrut{\rule{0pt}{2.6ex}}         
\newcommand\Bstrut{\rule[-0.9ex]{0pt}{0pt}}   

\usepackage{color,soul}
\usepackage[colorinlistoftodos,prependcaption]{todonotes}
\usepackage{algorithm}
\usepackage[noend]{algpseudocode}

\algnewcommand\algorithmicinput{\textbf{Input:}}
\algnewcommand\INPUT{\item[\algorithmicinput]}
\algnewcommand\algorithmicoutput{\textbf{Output:}}
\algnewcommand\OUTPUT{\item[\algorithmicoutput]}

\algnewcommand\algorithmicforeach{\textbf{for each}}
\algdef{S}[FOR]{ForEach}[1]{\algorithmicforeach\ #1\ \algorithmicdo}

\makeatletter
\renewcommand{\ALG@beginalgorithmic}{\small}
\makeatother
\algrenewcommand\alglinenumber[1]{\small #1:}

\usepackage{footnote}
\usepackage{url}
\usepackage{flushend}

\title{\LARGE \bf
An evolutionary algorithm for online, resource constrained, multi-vehicle sensing mission planning
}

\author{Nikolaos Tsiogkas$^{1}$ and David M. Lane$^{1}$%
\thanks{$^{1}$ Ocean Systems Lab, Heriot Watt University, Edinburgh UK {\tt\small \{nt95, d.m.lane\}@hw.ac.uk}}%
}

\begin{document}

\bstctlcite{IEEEexample:BSTcontrol}

\maketitle
\thispagestyle{empty}
\pagestyle{empty}

\begin{abstract}

Mobile robotic platforms are an indispensable tool for various scientific and industrial applications. Robots are used to undertake missions whose execution is constrained by various factors, such as the allocated time or their remaining energy. Existing solutions for resource constrained multi-robot sensing mission planning provide optimal plans at a prohibitive computational complexity for online application \cite{yu2014, yu2016, Tsiogkas2016}. A heuristic approach exists for an online, resource constrained sensing mission planning for a single vehicle \cite{Tsiogkas2017}. This work proposes a \emph{Genetic Algorithm} (GA) based heuristic for the \emph{Correlated Team Orienteering Problem} (CTOP) that is used for planning sensing and monitoring missions for robotic teams that operate under resource constraints. The heuristic is compared against optimal \emph{Mixed Integer Quadratic Programming} (MIQP) solutions. Results show that the quality of the heuristic solution is at the worst case equal to the 5\% optimal solution. The heuristic solution proves to be at least 300 times more time efficient in the worst tested case. The GA heuristic execution required in the worst case less than a second making it suitable for online execution.

\end{abstract}

\section{Introduction}
\label{sec:intro}

Robotic technology has significantly advanced during the past years. The maturing process of the scientific breakthroughs, combined with engineering effort, has led to reliable and affordable systems, that are becoming an indispensable assistant to science and industry. Some examples include space exploration \cite{Τruszkowski2006}, underwater archaeology \cite{Allotta2015} and search and rescue \cite{Tomic2012}.

In many cases the robot operation is constrained by the nature of the problem the robot is solving. For example, in emergency situations time is of essence. It is required to gather the maximum amount of information in a limited amount of time so that a response plan can be devised. In other cases the dynamic nature of the environment affects the energy requirements, leaving the vehicle with less energy to complete the required mission. For example, in a highly tidal environment the developed currents are strong and have an effect on the mission execution energy and time \cite{Willcox1996}. In such cases it is required to maximise the sensing outcome with limited energy by being able to replan a maximising trajectory online.

This paper focuses on sensing missions where specific resource constraints need to be respected. The mission requires a team  of vehicles to sample a user defined area for scalar field estimation. 
A scalar field is used to represent the physical quantities in space. Its importance is presented in \cite{La2015}, where phenomena such as algal blooms, oil spills and chemical pollution are studied. These phenomena are dynamic and they need to be monitored fast and precisely in order to be tackled. An example can be seen in \cite{Kinsey2011} with the use of Autonomous Underwater vehicles for monitoring in the Deepwater Horizon oil spill. For these cases \cite{yu2014, yu2016} presented a variant of the \emph{Orienteering Problem} (OP) for persistent monitoring tasks using \emph{Unmanned Aerial Vehicles} (UAVs). The presented approach takes into account the correlation of information among the sampling points, while respecting some budget constraints. The problem was named the \emph{Correlated Orienteering Problem} (COP) and \emph{Mixed Integer Quadratic Programming} (MIQP) formulations were provided for a single (COP) and multiple vehicles (CTOP). The method was able to generate paths for one or more vehicles that were optimal for the monitoring task.

In \cite{Tsiogkas2016} the COP method was adapted for an underwater sensing scenario. Results showed its potential over the standard OP for field estimation, but its computational complexity made it prohibitive for online use. To overcome that, in \cite{Tsiogkas2017}, a GA based heuristic was presented. The presented heuristic was able to generate paths for a single vehicle that were close to optimal, with low computational cost, making it suitable to be used online.

The work presented in this paper is focused on heuristic solutions for the multi-vehicle case of the COP.
It provides a GA based heuristic method that aims to generate close to optimal paths for the whole team, while keeping the time complexity low. This would allow plans to be regenerated online in case of a change in the mission parameters. For example, if some constraint changes, or in case of a vehicle failure, a new plan can be generated for the team, guaranteeing a close to optimal execution any time.

The rest of the paper is organised as follows. Section \ref{sec:prev} presents a brief review on previous work in the field of resource constrained optimisation. In section \ref{sec:methods} the CTOP and the proposed heuristic solution are detailed. Section \ref{sec:results} presents the results obtained through simulated experiments. Finally, in section \ref{sec:conclusion} the paper concludes discussing potential future paths for research.
\section{Previous work}
\label{sec:prev}
In the literature various approaches to maximise the utility gathered from an agent that operates under some resource constraint can be found. A formal definition of the problem was first presented in \cite{Laporte1990} as the selective travelling salesman problem. The work of \cite{Chao1996} coined the term \emph{orienteering problem} (OP), as the same problem describes the sport of orienteering. Applications of the problem to transportation and logistics systems are presented in \cite{Vansteenwegen2011, Gunawan2016}.


Extending the classic OP to multiple agents is mentioned in the literature as the \emph{team orienteering problem} (TOP). Over the years multiple heuristic solutions have been presented. In \cite{Tang2005} a tabu search heuristic is used. Local search methods are presented in \cite{Vansteenwegen2009a} and \cite{Vansteenwegen2009b}. The work of \cite{Ke2008} presents an ACO approach, while \cite{Lin2012} uses simulated annealing to solve the problem. Variable neighbourhood search is used in \cite{Labadie2012}. In \cite{Aghezzaf2014} \emph{variable neighbourhood search} (VNS) is used to tackle the Capacitated TOP with time windows. A path relinking approach for the TOP is presented in \cite{Souffriau2010}. In \cite{Labadie2011} the TOP with time windows is solved using an evolutionary local search algorithm.

The tasks of sampling or information gathering have been modelled as instances of the OP in various works. In \cite{singh2009} it is applied to lake and river monitoring. Approximation methods of a submodular OP are used to maximise the information gathering using autonomous surface vehicles. In \cite{binney2010} a recursive greedy algorithm is used to solve the additional information gathering problem in a sensor network. The work of \cite{heng2015} found the submodular OP inapplicable in real-time tasks. It proposes the usage of a linear approximation of the submodular OP for area coverage using a micro-UAV. A branch-and-bound solution to the OP is used in \cite{Binney2012} for scalar field estimation using Gaussian processes. In \cite{frolov2014} a GA based heuristic is used to solve an OP for marine sampling. In \cite{tokekar2013} an agricultural monitoring application is shown using multiple robots for sensing. The work of \cite{Azi2010} presents a vehicle routing problem with time windows variant that can't serve all the customers so it is forced to select customers based on revenue. That problem formulation is directly related with the TOP. The problem is solved using a branch-and-price approach. In \cite{Popovic2017} an evolutionary strategy is used for online terrain classification using unmanned aerial vehicles. Informative path planning under disturbances is studied in the work of \cite{Ma2016}. They provide a dynamic programming solution to the addressed path planning problem. In \cite{Pvenivcka2017DOP, Pvenivcka2017DNOP} they present a VNS approach for Dubins vehicles to the OP and COP respectively. Their approach successfully computes optimised paths, but the computational cost is prohibitive for online application. The work of \cite{Arora2017} provides a randomised algorithm for the informative path planning for a single vehicle. The computational time is allowing it to be applied online. To the best of our knowledge there is no prior work to tackle the online correlated informative path planning under budget constraints for multiple vehicles.

\section{Methods}
\label{sec:methods}
In this section the proposed solution and the methods that it consists of are presented in detail. Subsection \ref{ssec:gp} briefly presents a method for scalar field estimation using Gaussian processes. In \ref{ssec:ctop} the MIQP formulation of the CTOP is presented. Subsection \ref{ssec:heuristic} presents the details of the proposed evolutionary heuristic.

\subsection{Scalar field estimation}
\label{ssec:gp}
As mentioned in section \ref{sec:intro} the mission requires a scalar field to be estimated. A method to model such a field based on a \emph{Gaussian process} (GP) is presented in \cite{Binney2012}. To achieve that, a grid of sampling points is used to approximate the field. The field values are modelled as a multivariate Gaussian distribution, where one dimension is used for each sampling point. The covariance is calculated using a kernel function, such as the square exponential kernel shown in \eqref{eq:covar}.

{ \small 
  \setlength{\abovedisplayskip}{6pt}
  \setlength{\belowdisplayskip}{\abovedisplayskip}
  \setlength{\abovedisplayshortskip}{0pt}
  \setlength{\belowdisplayshortskip}{3pt}
\begin{align}
  &k(x_{i},x_{j})=exp(-\frac{\|x_{i}-x_{j}\|}{2l^{2}})
  \label{eq:covar}
\end{align}
}%

The kernel is used to model the drop in correlation of two points based on their distance. The parameter $l$ governs that behaviour and can be estimated from previously gathered data. In general, the distance in which two points can be correlated is larger, when $l$ grows larger. This is the correlation taken into account in the aforementioned CTOP formulation.

\subsection{Correlated Team Orienteering Problem}
\label{ssec:ctop}
The CTOP, which was first defined in \cite{yu2014, yu2016}, maximises the following objective function: 

{ \small 
  \setlength{\abovedisplayskip}{6pt}
  \setlength{\belowdisplayskip}{\abovedisplayskip}
  \setlength{\abovedisplayshortskip}{0pt}
  \setlength{\belowdisplayshortskip}{3pt}
\begin{align}
&\sum_{i\in V}(r_{i}x_{i}+\sum_{v_{j}\in N_{i}}r_{j}w_{ij}x_{i}(x_{i}-x_{j})) \label{eq:cop_obj}
\end{align}
}%

For any given solution the objective function represents the sum of the reward of each of the visited vertices in that solution. All the vertices are included in set $V$. The variable $r_{i}$ represents the reward for visiting vertex $i$ in a solution. To encode that a vertex $i$ is visited in a given solution, the binary variable $x_{i}$ is used. In addition to the reward received for visiting a vertex $i$, extra reward is obtained from its neighbourhood $N_{i}$. This reward is calculated by summing the reward of each unvisited neighbour $j$ multiplied by a weight $w_{ij}$. To ensure that only reward from unvisited vertices will be added, the quadratic term $x_{i}(x_{i}-x_{j})$ is used.

To enforce that $m$ vehicles start and finish in user defined points, constraints \eqref{eq:cop_cons_1} and \eqref{eq:cop_cons_2} are used.

{ \small 
  \setlength{\abovedisplayskip}{6pt}
  \setlength{\belowdisplayskip}{\abovedisplayskip}
  \setlength{\abovedisplayshortskip}{0pt}
  \setlength{\belowdisplayshortskip}{3pt}
\begin{align}
&\sum_{i\in V\setminus\{s\}} x_{isk} = \sum_{i\in V\setminus\{f\}} x_{fik} = 0, &\forall 1 \leq k \leq m \label{eq:cop_cons_1}\\
&\sum_{i\in V\setminus\{s\}} x_{sik} = \sum_{i\in V\setminus\{f\}} x_{ifk} = m, &\forall 1 \leq k \leq m \label{eq:cop_cons_2}
\end{align}
}%

In this formulation a binary variable $x_{ijk}$ is showing that a path exists from vertex $i$ to vertex $j$ in the path of vehicle $k$. Specifically, $x_{sik}$ shows that any vertex $i$ is visited right after the starting vertex $s$ for the path of vehicle $k$. Likewise, variable $x_{ifk}$ defines that the finishing vertex is after vertex $i$. For all the other vertices in the solution constraints \eqref{eq:cop_cons_3}-\eqref{eq:cop_cons_7} are applied.

{ \small 
  \setlength{\abovedisplayskip}{6pt}
  \setlength{\belowdisplayskip}{\abovedisplayskip}
  \setlength{\abovedisplayshortskip}{0pt}
  \setlength{\belowdisplayshortskip}{3pt}
\begin{align}
&\sum_{j\in V\setminus\{sf\}} x_{ijk} \leq 1, &\forall i\in V\setminus{s,f}, 1 \leq k \leq m \label{eq:cop_cons_3}\\
&\sum_{j\in V\setminus\{sf\}} x_{jik} \leq 1, &\forall i\in V\setminus{s,f}, 1 \leq k \leq m \label{eq:cop_cons_4}
\end{align}
}%

In the solution, each vertex is visited at most once, as it is ensured from constraints \eqref{eq:cop_cons_3} and \eqref{eq:cop_cons_4}. This behaviour is enforced by allowing at most one path entering and one leaving in each of the vertices in V, other than the start and finish. This happens for all the $m$ paths.

{ \small 
  \setlength{\abovedisplayskip}{6pt}
  \setlength{\belowdisplayskip}{\abovedisplayskip}
  \setlength{\abovedisplayshortskip}{0pt}
  \setlength{\belowdisplayshortskip}{3pt}
\begin{align}
&\sum_{k=1}^{m}\sum_{j\in V\setminus\{sf\}} x_{ijk} = x_{i} \leq 1, &\forall i\in V\setminus{f} \label{eq:cop_cons_5}\\
&\sum_{k=1}^{m}\sum_{j\in V\setminus\{sf\}} x_{jik} = x_{i} \leq 1, &\forall i\in V\setminus{s} \label{eq:cop_cons_6}\\
&\sum_{\mathclap{\j\in V\setminus\{sf\}}} x_{ijk}=\sum_{\mathclap{j\in V\setminus\{sf\}}} x_{jik}, &\forall i\in V, 1 \leq k \leq m \label{eq:cop_cons_7} 
\end{align}
}%

Constraints \eqref{eq:cop_cons_5} and \eqref{eq:cop_cons_6} make sure that if a vertex $j$ is visited in a solution, a path that connects $j$ with a next vertex exists. Constraint \eqref{eq:cop_cons_7} prevents a vertex having a path from one vehicle entering and the path of a different vehicle exiting it. The maximum amount of resources used is enforced by constraint \eqref{eq:cop_cons_8}.

{ \small 
  \setlength{\abovedisplayskip}{6pt}
  \setlength{\belowdisplayskip}{\abovedisplayskip}
  \setlength{\abovedisplayshortskip}{0pt}
  \setlength{\belowdisplayshortskip}{3pt}
\begin{align}
&\sum_{i\in V}\sum_{j\in V} (c_{ij} + c_{i}) x_{ijk} \leq C_{k}, &\forall 1 \leq k \leq m \label{eq:cop_cons_8}
\end{align}
}%

Term $c_{i}$ represents a fixed cost of performing sensing in vertex $i$. To that it is added the cost of travelling from vertex $i$ to the next vertex $c_{ij}$. Then the cost is multiplied by the binary variable $x_{ijk}$, used to denote the existence of a path between $i$ and $j$ in path of robot $k$. The sum of all these costs must be less or equal to the maximum allowed resource usage for robot $k$. It must be noted that the fixed cost of operations of the user defined start and finish vertices is zero.

{ \small 
  \setlength{\abovedisplayskip}{6pt}
  \setlength{\belowdisplayskip}{\abovedisplayskip}
  \setlength{\abovedisplayshortskip}{0pt}
  \setlength{\belowdisplayshortskip}{3pt}
\begin{align}
\begin{split}
& u_{ik} - u_{jk} + 1 \leq (\left\vert{V}\right\vert - 1)(1 - x_{ijk})\\ &\forall i,j \in V, i\neq j, \forall 1 \leq k \leq m 
\end{split}\label{eq:cop_cons_9}\\
\begin{split}
& 0 \leq u_{ik} \leq \left\vert{V}\right\vert \\ &\forall i\in V, \forall 1 \leq k \leq m
\end{split} \label{eq:cop_cons_10}
\end{align}
}%

Finally,  constraints \eqref{eq:cop_cons_9} and \eqref{eq:cop_cons_10} are subtour elimination constraints, allowing only a single tour to be generated for each robot.

\subsection{Genetic Algorithm heuristic}
\label{ssec:heuristic}

Genetic algorithms are one of the main meta-heuristics used for optimisation. They try to mimic the evolution process occurring in nature by using a population of chromosomes that undergo processes of natural selection, reproduction and mutation as described in \cite{Mitchell1998}. The genetic algorithm presented by this paper uses the chromosome encoding as presented in \cite{Singh2009ga}. It is composed by a set vectors, or genes, each one of which is representing a path for a single robot. In each vector the indices of the vertices to be visited by that robot is stored. The whole chromosome, therefore, represents a candidate CTOP solution. An outline of the procedure used during the optimisation can be seen in algorithm \ref{alg:ga}. It can be seen that the optimisation procedure is separated in two distinct phases. The first phase is responsible for the population generation, while the second phase repeatedly applies the operations of selection, crossover and mutation to optimise the population. 
\begin{algorithm}[]
\caption{Genetic Algorithm for the Correlated Team Orienteering Problem}
\label{alg:ga}
\begin{algorithmic}[1]
\INPUT $TravelCosts, Rewards, MaxCosts, Start, Finish$
\OUTPUT $FittestChromosome$
\State $\Call{InitialisePopulation}{}$
\While{$gen \le MaxGen$ or $FittestChromosome$ stable for 10 generations}
	\State $\Call{SelectNewPopulation}{}$
	\State $\Call{Crossover}{}$
	\State $\Call{Mutate}{}$
\EndWhile
\State \textbf{select} $FittestChromosome$
\State \Return $FittestChromosome$
\end{algorithmic}
\end{algorithm}

In the scope of this paper two different chromosome generation methods are presented and compared. The difference is based on the way the genes, composing the chromosome, are generated. The first method is a random method, similar to the one presented in \cite{Tsiogkas2017}. It can be seen in algorithm \ref{alg:random}. 

\begin{algorithm}[]
\caption{Random Gene Generation}
\label{alg:random}
\begin{algorithmic}[1]
\INPUT $TravelCosts, Rewards, MaxCost,$ $Start, Finish, AvailVertices$
\OUTPUT $Gene, AvailVertices$
\State $\Call{Insert}{Gene, Start}$
\State $Done$ $\gets$ $False$
\While{$!Done$ \textbf{and} $AvailVertices \neq \varnothing$}
	\State $v \gets $ \Call{SelectRandom}{$AvailVertices$}
	\If{$\Call{CostInsert}{Gene,v} \leq MaxCost$}
		\State $\Call{Insert}{Gene,v}$
		\State $AvailVertices \gets AvailVertices \setminus v$
	\Else
		\State $Done \gets True$
	\EndIf
\EndWhile
\State $\Call{Insert}{Gene, Finish}$
\State \Return $Gene, AvailVertices$
\end{algorithmic}
\end{algorithm}

In the random generation method genes are created by randomly inserting vertices until the maximum cost is reached or until there are no more free vertices to insert. To generate a chromosome this process is repeated a number of times, equal to the number of robots. In the chromosome generation process, the set of available vertices is shared among the genes, so that if a vertex is picked up by a robot, it cannot be assigned to any other robot. 

The second gene generation method is named \emph{Nearest Neighbour Randomised Adaptive Search Procedure} (NN-RASP). It is inspired by the \emph{Greedy Randomised Adaptive Search Procedure} (GRASP) that is used in \cite{Souffriau2010} to provide a heuristic solution to the TOP. This procedure is taking into consideration the peculiarities of the CTOP problem, regarding the correlation of information. It  and can be seen in algorithm \ref{alg:nnrasp}.

\begin{algorithm}[]
\caption{Nearest Neighbour Randomised Adaptive Search Procedure}
\label{alg:nnrasp}
\begin{algorithmic}[1]
\INPUT $TravelCosts, Rewards, MaxCost, Start, Finish$
\OUTPUT $Gene$
\State $\Call{Insert}{Gene, Start}$
\State $v \gets \Call{GetLast}{Gene}$
\State $N \gets \Call{GetNeighbours}{v}$
\While{$\Call{GetSize}{N} > 0$}
	\ForEach{$n \in N$}
		\State $\Call{GetDistanceWeight}{n}$
		\State $\Call{GetNeighbourWeight}{n}$
		\State $\Call{GetCombinedWeight}{n}$
	\EndFor
	\State $nv \gets \Call{GetWeightedRandom}{N}$
	\If{$\Call{CostInsert}{Gene, nv} <= MaxCost$}
		\State $\Call{Insert}{Gene,nv}$
		\State $v \gets \Call{GetLast}{Gene}$
		\State $N \gets \Call{GetNeighbours}{v}$
	\Else
		\State $N \gets \{\varnothing\}$
	\EndIf
\EndWhile
\State $\Call{Insert}{Gene, Finish}$
\State \Return $Gene$
\end{algorithmic}
\end{algorithm}

This method is based on progressively constructing a path by choosing one of the neighbours of the last inserted vertex in the path. The neighbour to be inserted is chosen based on a categorical distribution where each one has a probability to be chosen. This probability is based on the distance of the neighbouring vertex from all other visited vertices in the solution and the number of free neighbours it has. The distance from other vertices is chosen as this can reduce overlapping of paths. The number of free neighbours is used as it gives a better reward from correlated information, as well as, it will allow the path to have more options to continue its construction. When a vertex is chosen it is attempted to be inserted in the gene. If the gene that is resulting is feasible the vertex is inserted and the search continues, else the search stops and the gene is returned. The chromosome generation is again performed by iteratively calling the gene generation methods until paths are generated for all the robots.

To create the population both chromosome generation procedures are repeated until the specified amount of chromosomes is generated. Following their generation, the chromosomes are evaluated according to algorithm \ref{alg:eval}.

\begin{algorithm}[]
\caption{Chromosome Evaluation}
\label{alg:eval}
\begin{algorithmic}[1]
\INPUT $TravelCosts, Rewards, MaxCost$
\OUTPUT $Fitness$
\ForEach{$g \in Genes$}
	\State $\Call{TwoOpt}{g}$
	\State $\Call{RemoveDuplicates}{g}$
	\State $GeneFit \gets \Call{GetReward}{g}^3/\Call{GetCost}{g}$
	\State $Fitness \gets Fitness + GeneFit$
\EndFor
\State \Return $Fitness$
\end{algorithmic}
\end{algorithm}

The fitness of each chromosome is defined as the sum of fitnesses of each gene composing it. Each gene goes through a \emph{2-Opt} operation to minimise its cost. This operation is a simple local search method that is based on swapping segments of a path in hope of reducing the path's cost. It is iteratively applied to all segment combinations and stops if no improvement can be achieved. It was first presented in \cite{Croes1958} as a method to solve the travelling salesman problem. Then any duplicate vertices that have been already visited by other genes are removed. Its fitness is calculated as the total reward to the third power divided by the gene cost as suggested in \cite{Karbowska2012} for the single agent orienteering problem.

After the population generation is completed, a repetitive process is applied. This process is applied until either the best chromosome is stable for 10 generations or a maximum amount of generations is reached. This process involves selecting a new population based on the previous one. A percentage of the old population is passed to the next generation without selection. This process is described in the literature as \emph{Elitism} and is known to speed up the performance and prevent loss of good solutions during the search process \cite{li2009}. The rest of the population is selected using \emph{Tournament selection}. This method starts with an empty population, then repetitively chooses $n$ random individuals from the previous population and adds the fittest to the new population, until the maximum amount of individuals is chosen.

Following the new population selection a crossover operation is performed to a percentage of the population. The operation is presented in algorithm \ref{alg:cx} and is a modified version of the one presented in \cite{Singh2009ga}. The crossover operation aims to diversify the population and explore more of the search space.

\begin{algorithm}[]
\caption{Chromosome Crossover}
\label{alg:cx}
\begin{algorithmic}[1]
\INPUT $Parent_{1}, Parent_{2}$
\OUTPUT $Child_{1}, Child_{2}$
\State $\Call{SortGenes}{Parent_{1}$}
\State $\Call{SortGenes}{Parent_{2}$}
\ForEach{$i \in {1,2}$}
	\State $tP_{1} \gets Parent_{1}$
	\State $tP_{2} \gets Parent_{2}$
	\While{$\Call{GetNumGenes}{Child_{i}} < numRobots}$
		\State $p$ $\gets$ \Call{SelectRandomParent}{$tP_{1}$, $tP_{2}$}
		\State $\Call{InsertBestGene}{Child_{i},p}$
		\State \Call{RemoveVertices}{$tP_{1}$, $tP_{2}$}
		\State \Call{EvaluateAndSortGenes}{$tP_{1}$, $tP_{2}$}
	\EndWhile
	\State $\Call{EvaluateChromosome}{Child_{i}}$
\EndFor
\State \Return $Child_{1}, Child_{2}$
\end{algorithmic}
\end{algorithm}

For this operation to take place, two chromosomes are randomly chosen as the parents of the crossover and they are replaced by the two produced offsprings. Initially the genes of each parent are sorted by fitness. Then, the two children are constructed iteratively. The first step of the iteration involves copies of the parents to be created. Next, as long as the number of genes in each child is less than the number of robots, the following process takes place. A parent is selected randomly and its best gene is removed and inserted to the child. Then, the vertices of this gene are removed from any other genes of the temporary parents. After that the genes of the parents are evaluated again and sorted based on their fitness. When the maximum amount of genes is inserted in the child, it is accordingly evaluated. After both offsprings are generated, they are returned in the place of their parents.

Continuing the population crossover, a percentage of the population is evolving by mutating. This aims at attempting to improve the fitness of each gene that composes the mutating chromosome and thus improve the fitness of the chromosome itself. The mutation process resembles the one presented in \cite{Tsiogkas2017} and can be seen in algorithm \ref{alg:mutate}.

\begin{algorithm}[]
\caption{Chromosome Mutation}
\label{alg:mutate}
\begin{algorithmic}[1]
\INPUT $Chromosome, MaxCost, NumMut, AddProb$
\OUTPUT $Chromosome$
\State $Genes \gets \Call{GetGenes}{Chromosome}$
\ForEach{$g \in Genes$}
	\For{$i \gets 1, NumMut$}
		\State $p \gets \Call{GetRandom}{0,1}$
		\If{$p \leq AddProb$}
			\If{$\Call{GetCost}{g} \geq 0.95*MaxCost$}
				\State $v \gets  \Call{SelectRandom}{g, 1}$
				\State $n \gets  \Call{GetMaxFreeNeighbour}{v}$
				\State $gn \gets \Call{Replace}{g,v,n}$
				\If{$\Call{Fitness}{g} \leq \Call{Fitness}{gn}$}
					\State $g \gets gn$
				\EndIf
			\Else
				\If{$FV \neq [\varnothing]$}
					\State $v \gets  \Call{SelectRandom}{FV, 1}$
					\State $idx \gets  \Call{FindBestInsertion}{g, v}$
					\State $\Call{Insert}{g,v,idx}$
				\EndIf
			\EndIf
		\Else
			\State $v \gets  \Call{GetMinLossVertex}{g}$
			\State $g \gets \Call{Remove}{g, v}$ 
		\EndIf
	\EndFor
\EndFor
\State $\Call{EvaluateChromosome}{Chromosome}$
\State \Return $Chromosome$
\end{algorithmic}
\end{algorithm}

For each gene in the chromosome a total number of $NumMut$ mutations will take place. In the case studied in this work $NumMut$ was set to $10$. A mutation can happen in two ways, by either adding a vertex in the gene, or by removing one. This is governed by the $AddProb$ probability and was set to be $0.9$. In the first case there are two behaviours depending on the cost of the gene that is mutated. If the cost is $95\%$ of the maximum cost, an improvement of the solution is attempted by swapping a vertex of the solution with one of its free neighbours. A random vertex in the solution is chosen and is iteratively swapped with all its free neighbours. The swap with the maximum improvement in the gene's fitness is kept as the performed mutation. In the second behaviour, a free vertex is randomly chosen. Then the best insertion place is found in the path. It is consequently inserted if it does not violate the maximum cost constraint. The best insertion is found using the heuristic value described in the GRASP method in \cite{Souffriau2010}. In the second case, where a removal is performed, the minimum loss vertex for this gene's fitness is found and removed. Likewise with the insertion, the minimum loss vertex is found using the same heuristic value. After all the $NumMut$ mutations happen for all the genes, the chromosome is evaluated using the method already described and is returned.

\section{Experimental Results}
\label{sec:results}
This section describes the experimental results obtained through simulations. The planning is required for a simulated sampling mission that is to be performed by a team of vehicles. The mission is composed by a set of \textit{sampling points}, arranged in a grid. The team is given specific starting and finishing points. Additionally, each vehicle is allowed to use a limited amount of budget to complete the mission. The overall goal of the team is to maximise the utility obtained without exceeding the budget limitations. 

The experiments are divided into two sections. In the first section a method from the literature is used to tune the different parameters that govern the performance of the heuristic algorithm. In the second section the tuned parameters are used and proposed heuristic is compared against the optimal MIQP solution. The two methods are compared based on the utility obtained by the team and the time it took to generate a plan. The grid was chosen to be 9 by 9, as choosing a larger would make it impractical to collect results with the exact method. To further limit the planning time of the exact MIQP method, it was stopped once the solution found was less than $5\%$ different from the optimal. These results are referred with the term $CTOP 5\%$. For the heuristic, given that it is a random process, 1000 experiments were run for each case and statistical results are presented.

All the experiments were run on a system having an Intel i5-7600 CPU, 16GB of RAM and running on Ubuntu 16.04. The MIQP instances were solved using the Gurobi 7.0.2 commercial solver \cite{gurobi}. The heuristic was coded in C++11 and compiled using g++ version 5.4.0. Source code for both approaches can be found online \footnote{\url{https://github.com/lounick/lwga}}$^{,}$\footnote{\url{https://github.com/lounick/task_scheduling}}.

\subsection{Algorithm Tuning}
As described in section \ref{sec:methods}, a GA uses a set of parameters to manage its behaviour. A tuning process is essential to find the best parameters for the algorithm. In the literature one can find methods for parameter tuning. In \cite{Eiben2011} the general framework for parameter tuning is presented and various methods are discussed. In the scope of this work a modified version of the work presented in \cite{Vevcek2016} was used.

The parameter tuning method that \cite{Vevcek2016} presents is based on an evolutionary approach, using chess rating for evaluation. Each chromosome in this method is represented by a vector of integer and real values representing the set of parameters. To ensure that the tuned parameters are generalising well, each parameter set is tested against a set of problems. For tuning the presented GA, twelve problem instances were created. They were created by altering the problem size, the vertex distribution, and the number of vehicles. Problem instances of 5x5, 7x7 and 9x9 were used. The vertices were distributed in a grid and a noisy grid form. The noisy grid was created based on the canonical grid and adding some uniform random noise to the position of each vertex. The number of vehicles were 3 and 5. The budget was set to $75\%$ of the maximum budget for all the problems. The rest of the approach is presented in algorithm \ref{alg:tune}.

\begin{algorithm}[]
\caption{CRS-Tuning algorithm}
\label{alg:tune}
\begin{algorithmic}[1]
\INPUT $Problems$
\OUTPUT $BestConfiguration$
\State $C \gets \Call{InitialisePopulation}{}$
\For{$trial \gets 1, MaxTrials$}
	\ForEach{$c \in C$}
		\ForEach{$p in Problems$}
			\For($g \in 1, NumGames$)
				\State $s$ $\gets$ $GA(c,p)$
				\State $S_{c,p,g}$ $\gets$ $s$
			\EndFor
		\EndFor
	\EndFor

	\State \Call{EvaluateConfigurations}{$C, S$}
	\State $parents$ $\gets$ \Call{SelectParents}{$C$}
	\State $newC$ $\gets$ \Call{GenerateNewPop}{parents}
	\State $C$ $\gets$ $newC$
\EndFor
\State \textbf{select} $BestConfiguration$
\State \Return $BestConfiguration$
\end{algorithmic}
\end{algorithm}

The first step of the tuning algorithm is to generate a population of chromosomes. Each chromosome is constructed by randomly choosing values for the parameters from a specified range. For the tuning procedure performed for this work, the chromosome population was composed by 100 individuals. Then the algorithm iteratively runs for $MaxTrials$. For this work $MaxTrials$ were set to ten. Initially, each configuration is used to run each problem $NumGames$ times and their scores are recorded. Here $NumGames$ was set to ten as well.

Following, the configurations are evaluated using the \emph{Glicko2} chess rating system \cite{Glickman2012}. The evaluation is performed by each configuration comparing its results against all the other configurations results and collecting $0, 0.5$ and $1$ points for loss, draw and win respectively. These points, as well as the opponents rating, are used to update the fitness of each configuration. In \cite{Vevcek2016} the evaluation is performed based only on the scores of the games. For the tuning performed in this paper a bonus of $0.1$ was given to the faster solution, while the slower solution got a penalty of $-0.1$. This was done to differentiate the cases where two solutions were equally good utility-wise but one was slower to compute than the other.

The next step involved selecting the parents that would generate the next generation of configurations. The ten best configurations were selected as parents as well as any other parents that performed close to the first ten. The choice was performed based on their chess rating intervals. The parents had a maximum size of half the total population. These new parents were a portion of the new population. The rest of the population was generated by applying uniform crossover and mutation operations with $50\%$ and $80\%$ percent respectively, as suggested by \cite{Vevcek2016}. Finally the best configuration was selected and returned.

The tuning method was run for both population generation methods presented in section \ref{sec:methods}. The resulted configurations as well as the parameter ranges can be seen in Table \ref{tab:param_tune}.

\begin{table}[]
\centering
\caption{Parameter ranges and tuned values for the two different methods}
\label{tab:param_tune}
\begin{tabular}{cccc}
\hline \hline
Parameter & Value Range & Random & NNRASP \Tstrut \\ \hline
Population Size & $[25,50,\dots,500]$ & $300$ & $250$ \Tstrut \\
Generations Number & $[5,10,\dots,50]$ & $40$ & $50$ \\
Tournament Size & $[3,4,\dots,10]$ & $6$ & $5$ \\
CX Probability & $[0.0,0.1,\dots,0.9]$ & $0.7$ & $0.9$ \\
Mutation Probability & $[0.0,0.1,\dots,0.9]$ & $0.6$ & $0.7$ \\
Elitist Percentage & $[0.01,0.02,\dots,0.20]$ & $0.19$ & $0.03$ \Bstrut \\
\hline
\end{tabular}
\end{table}
\subsection{Results}

\begin{figure}
    \centering
    \begin{subfigure}[b]{0.49\columnwidth}
        \centering
        \includegraphics[width=\textwidth]{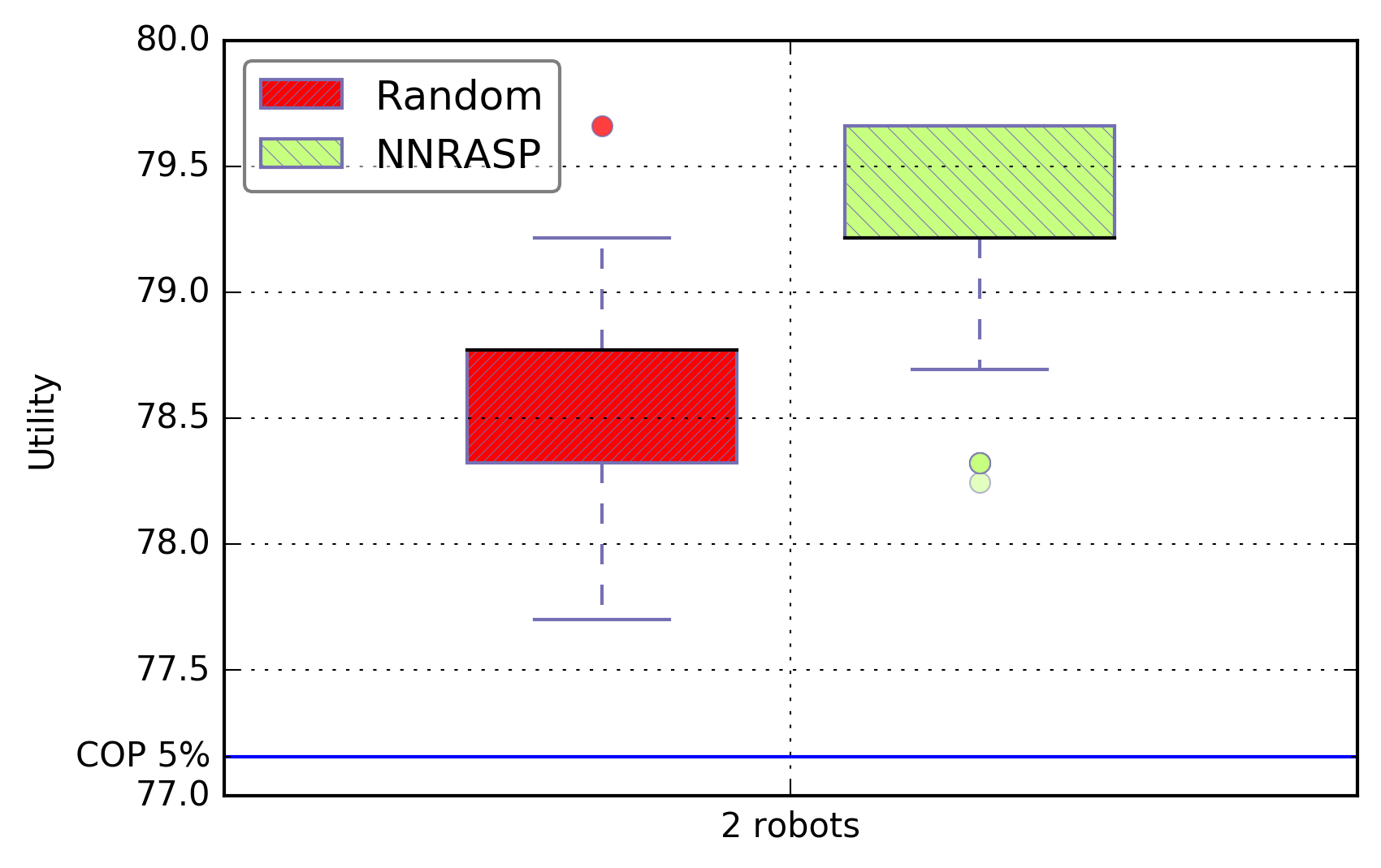}
    \end{subfigure}
    \begin{subfigure}[b]{0.49\columnwidth}  
        \centering 
        \includegraphics[width=\textwidth]{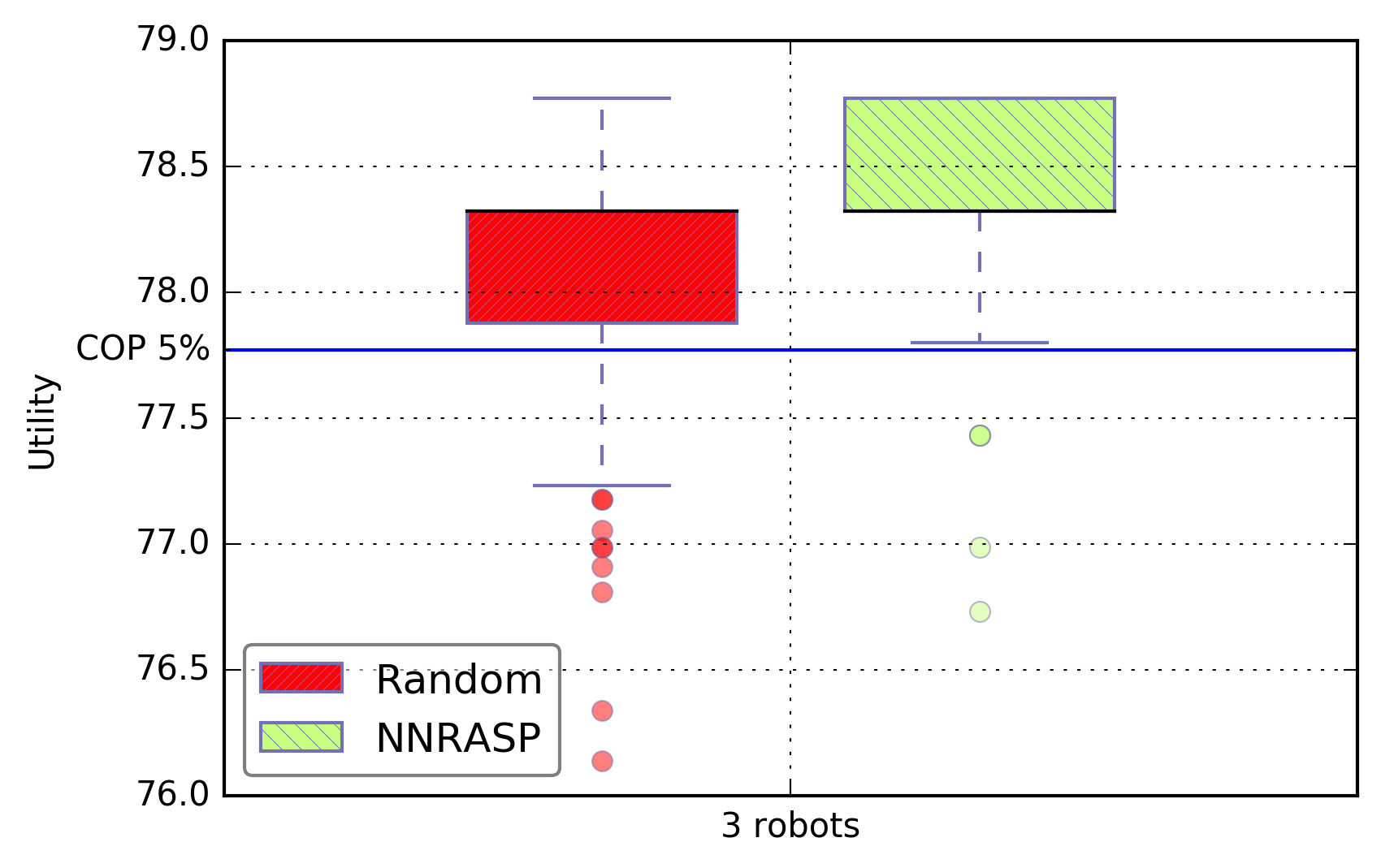}
    \end{subfigure}
    
    \begin{subfigure}[b]{0.49\columnwidth}   
        \centering 
        \includegraphics[width=\textwidth]{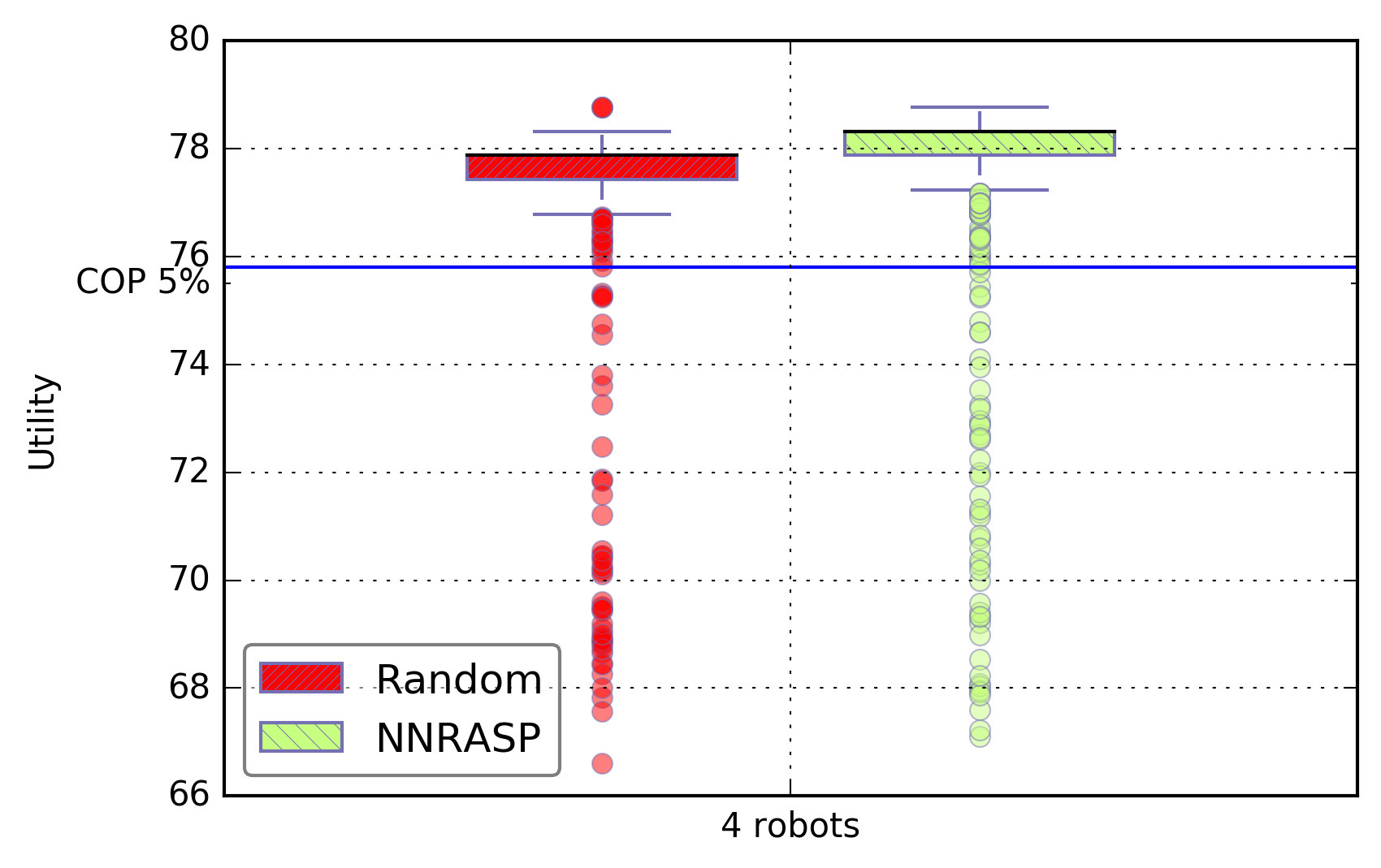}
    \end{subfigure}
    \begin{subfigure}[b]{0.49\columnwidth}   
        \centering 
        \includegraphics[width=\textwidth]{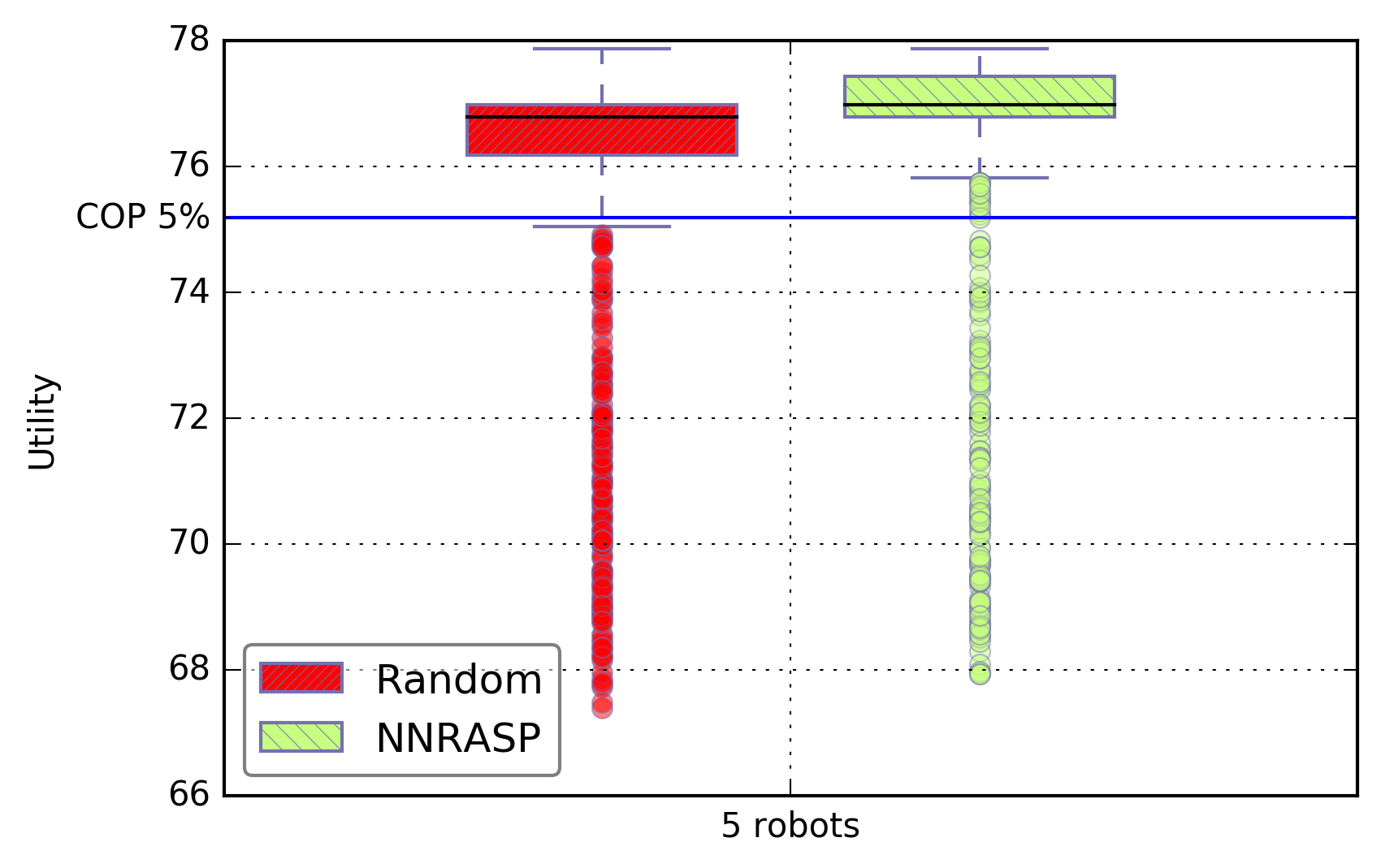}
    \end{subfigure}
    \caption{Utility box plots for different numbers of robots. For lower number of robots a sophisticated population initialisation method tends to give better results. As the number goes higher the results come closer with a slight superiority of the NNRASP method. The blue line represents the $5\%$ gap optimal solution obtained by solving the MIQP problem.} 
    \label{fig:util_num_robots}
\end{figure}

The performance of the heuristic method is compared against the optimal solution in two different ways. The first way examines how well the heuristic method performs for different numbers of robots. For that the maximum budget was allocated to each robot. The maximum budget was calculated by taking the maximum budget required for a single robot to complete the mission and dividing it by the number of robots. The results can be seen in figure  \ref{fig:util_num_robots} and Table \ref{tab:miqp_time_vehicle_scale}.


Figure \ref{fig:util_num_robots} presents statistical results regarding the utility for each size of the robotic team using the GA based planning. It is compared against a solution produced by solving the MIQP problem using the Gurobi solver. As it was already mentioned, the solver was stopped once the found solution was guaranteed to be at most $5\%$ worse than optimal. This solution is represented by the horizontal blue line. It can be seen that the GA heuristic with the NNRASP generation method outperforms the MIQP solution in all the cases. The random generation method falls a bit behind, with part of the results produced being worse than the $5\%$ optimal.

\begin{table}[]
\centering
\caption{Average time(s) for different algorithms and vehicle numbers}
\label{tab:miqp_time_vehicle_scale}
\begin{tabular}{lllll}
\hline \hline
\multirow{2}{*}{Algorithm} & \multicolumn{4}{c}{Number of Robots} \Tstrut \\
& 2 & 3 & 4 & 5  \Bstrut \\ \hline
CTOP 5\% &396.36 &695.60 &2830.41 &1474.18 \Tstrut \\
GA-Random & \textbf{0.787703} & \textbf{0.729135} & \textbf{0.666236} & \textbf{0.623074} \\
St.Dev. & 0.087695 & 0.0738167 & 0.0602966 & 0.0579891 \\
GA-NNRASP & 0.792475 & 0.844305 & 0.817855 & 0.826688 \\
St.Dev. & 0.106818 & 0.120596 & 0.112062 & 0.106136 \Bstrut \\
\hline
\end{tabular}
\end{table}

\begin{figure*}
  \centering
  \begin{subfigure}{.66\columnwidth}
    \centering
    \includegraphics[width=1.05\columnwidth, keepaspectratio]{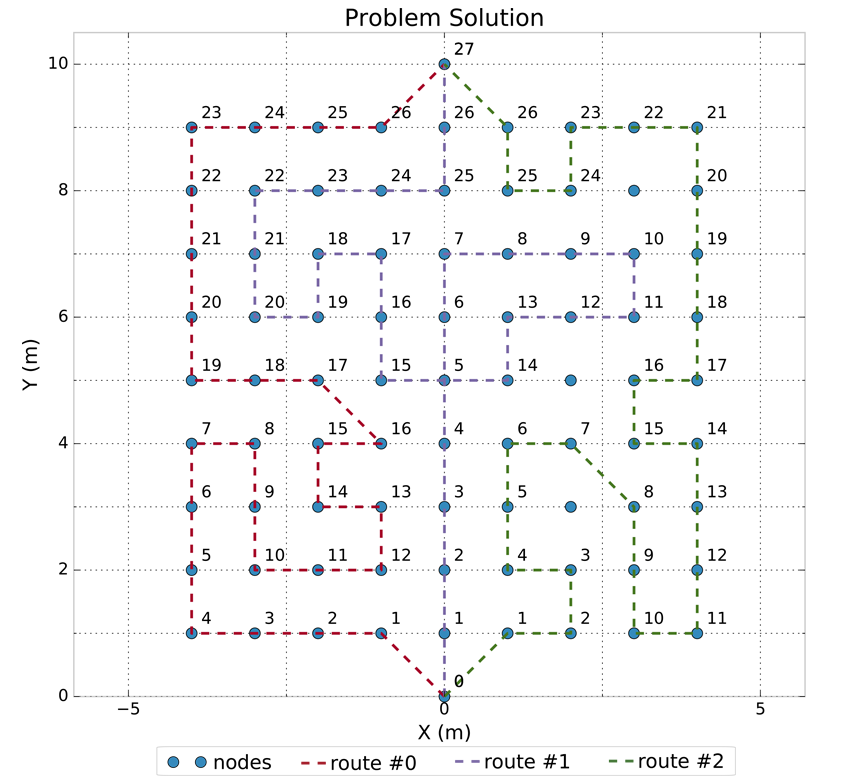}
    \caption{$1\%$ optimal MIQP solution for three vehicles. The gathered utility is 79.662 and the time to calculate is 5847.81 seconds.}
    \label{fig:miqp-3-optimal}
  \end{subfigure}\hfill
  \begin{subfigure}{.66\columnwidth}
    \centering
    \includegraphics[width=1.05\columnwidth, keepaspectratio]{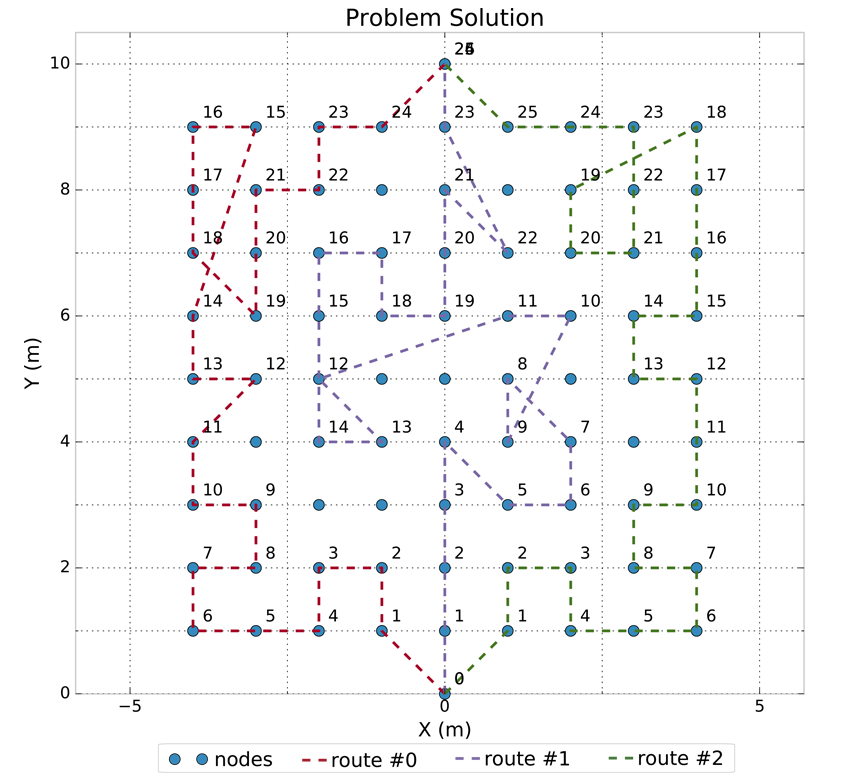}
    \caption{$5\%$ optimal MIQP solution for three vehicles. The gathered utility is 76.433 and the time to calculate is 690.74 seconds.}
    \label{fig:miqp-3-gap}
  \end{subfigure}\hfill
  \begin{subfigure}{.66\columnwidth}
    \centering
    \includegraphics[width=1.05\columnwidth, keepaspectratio]{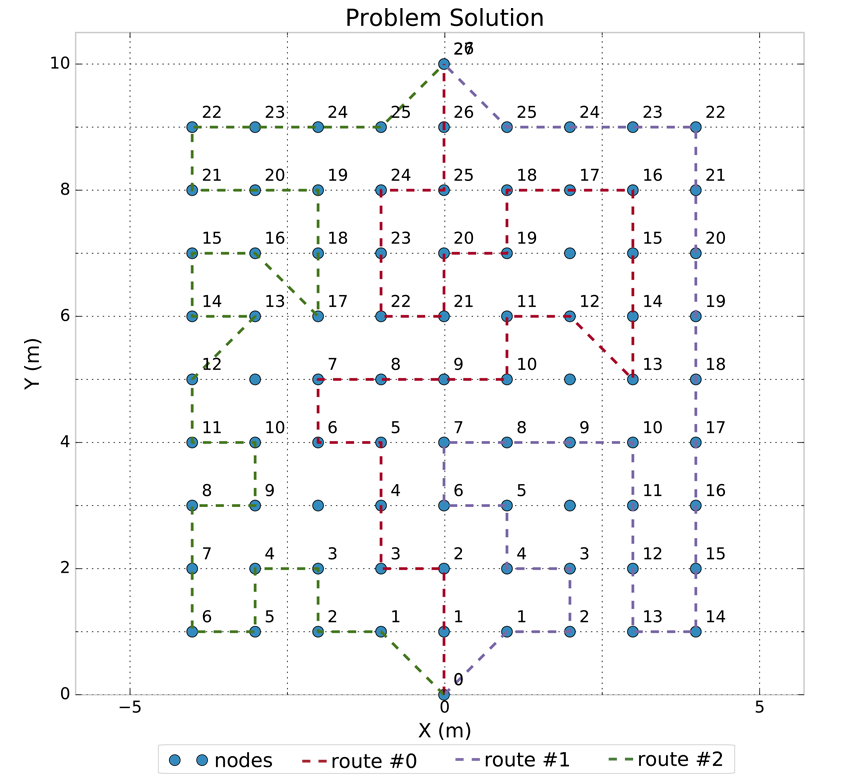}
    \caption{GA heuristic solution for three vehicles. The gathered utility is 78.770 and the time to calculate is 0.8597 seconds.}
    \label{fig:miqp-3-heuristic}
  \end{subfigure}
  \caption{Comparison of the paths generated for three vehicles having full budget.}
  \label{fig:full-energy-execution}
\end{figure*}

Comparing the average planning time the GA heuristic heavily outperforms the MIQP exact method, as it can be seen in Table \ref{tab:miqp_time_vehicle_scale}. It can be seen that the time required for the optimal solution is at the best case 500 times higher. The GA-Random method performed slightly better than the GA-NNRASP. Example paths for a three robot team can be seen in figures \ref{fig:miqp-3-optimal}, \ref{fig:miqp-3-gap} and \ref{fig:miqp-3-heuristic}. The first figure shows the $1\%$ optimal path calculated by the MIQP solver, while the second is a $5\%$ optimal solution used for comparison with the GA. The heuristic solution can be seen in the third image. It is obvious that the solution quality of the GA is far better than the $5\%$ optimal.

%

\begin{figure}
    \centering
    \begin{subfigure}[b]{0.49\columnwidth}
        \centering
        \includegraphics[width=\textwidth]{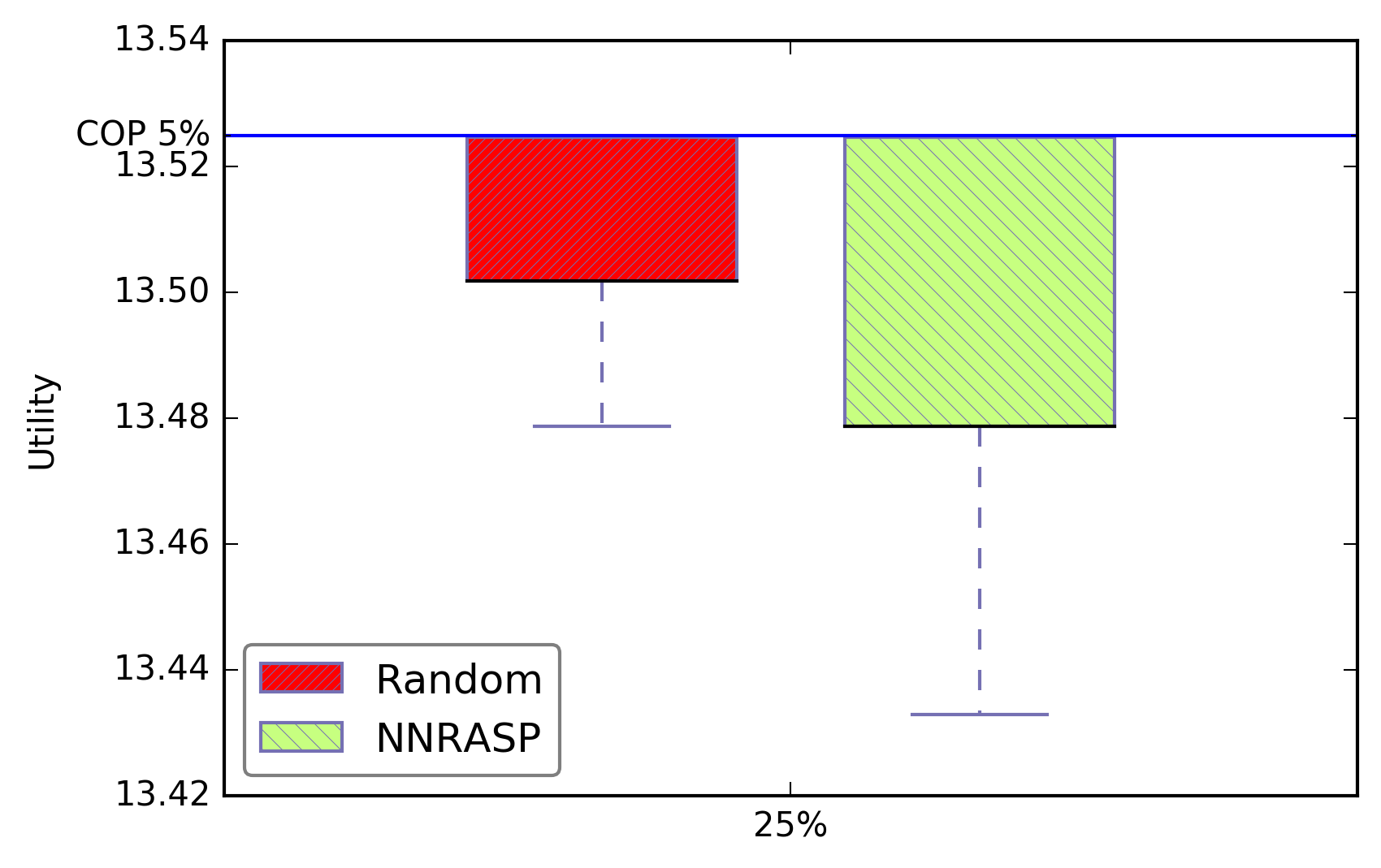}
    \end{subfigure}
    \begin{subfigure}[b]{0.49\columnwidth}  
        \centering 
        \includegraphics[width=\textwidth]{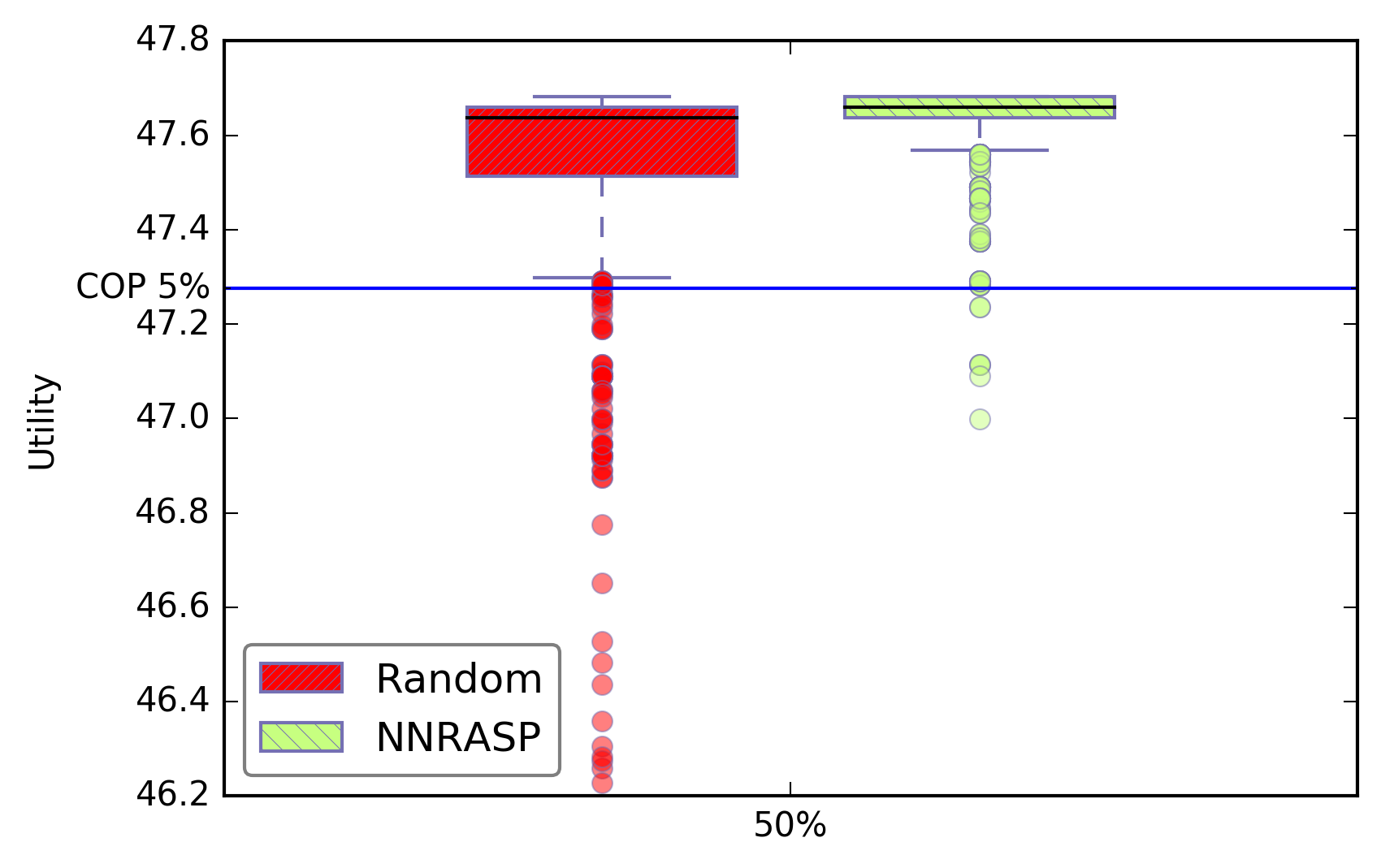}
    \end{subfigure}
    
    \begin{subfigure}[b]{0.49\columnwidth}   
        \centering 
        \includegraphics[width=\textwidth]{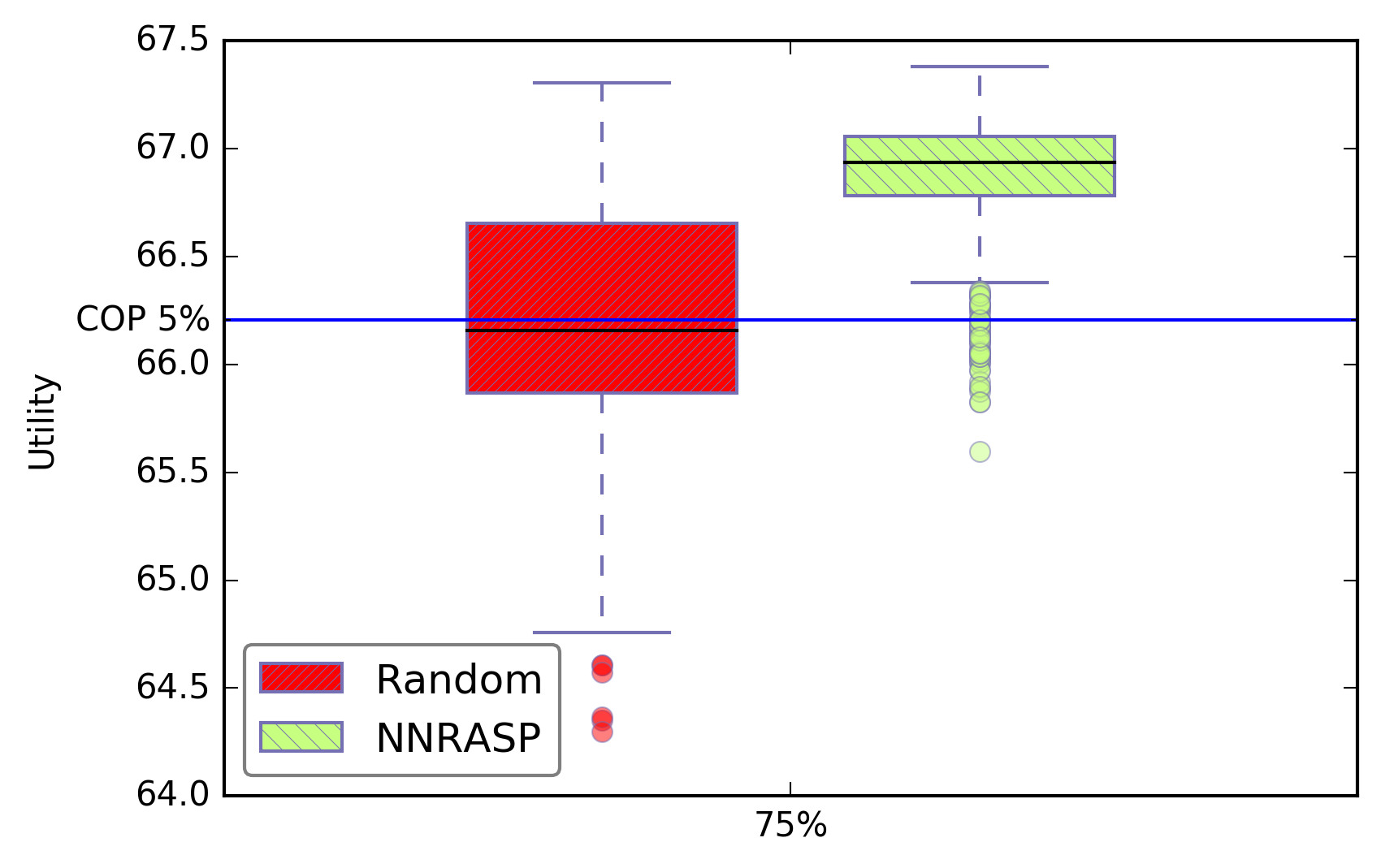}
    \end{subfigure}
    \begin{subfigure}[b]{0.49\columnwidth}   
        \centering 
        \includegraphics[width=\textwidth]{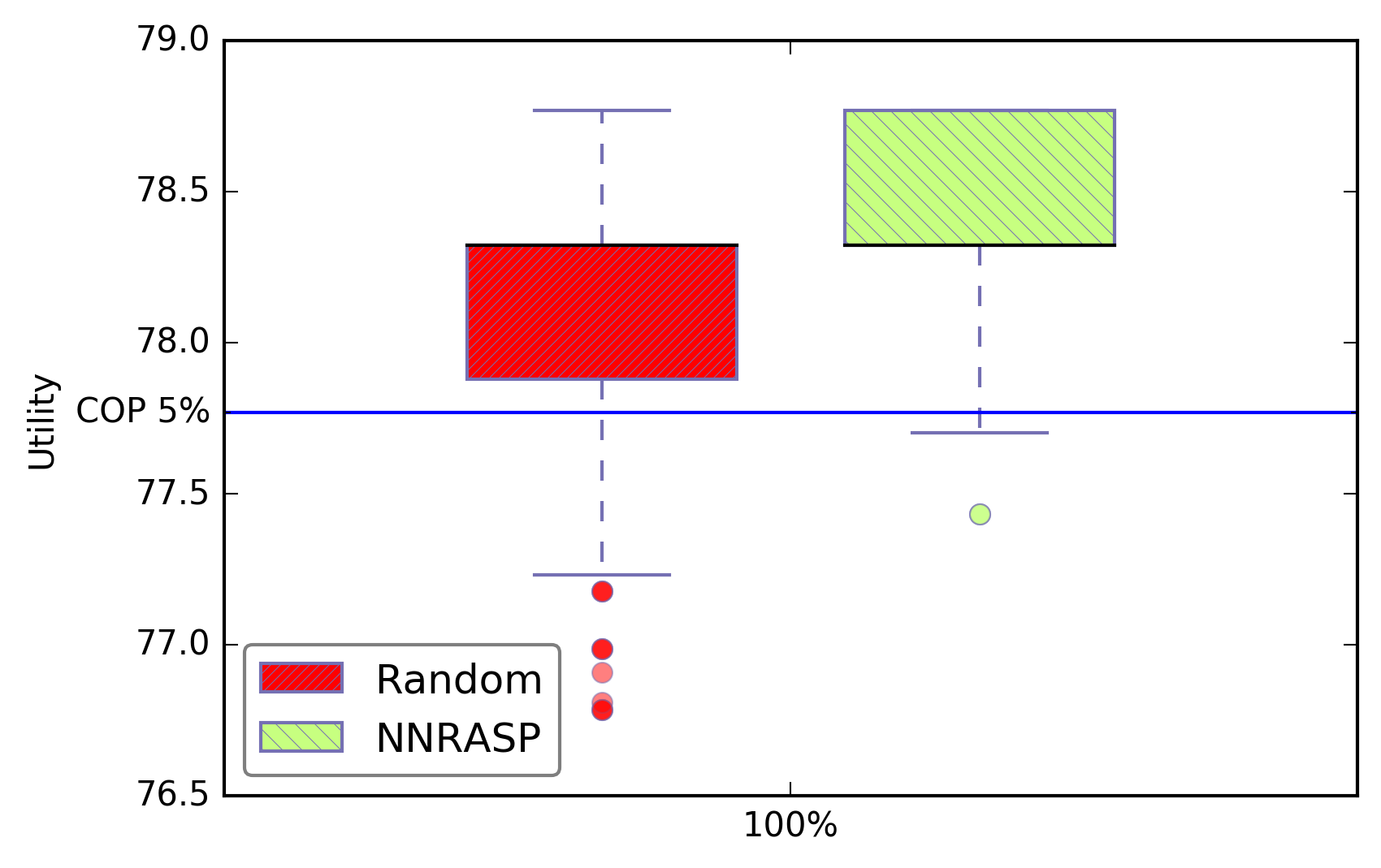}
    \end{subfigure}
    \caption{Utility box plots for three robots having different budget. For lower budgets a sophisticated population initialisation performs equally good with the random one. As the budget goes higher the results show the benefits of the NNRASP method. The blue line represents the $5\%$ gap optimal solution obtained by solving the MIQP problem.} 
    \label{fig:util_en}
\end{figure}

The second performance metric examines how the heuristic method performs in cases where the budget is limited. For this the number of vehicles is chosen to be three and is kept static. Then the budget of each vehicle is reduced by a specific percentage. Figure \ref{fig:util_en} and table \ref{tab:miqp_limited_energy_time} present the results for this metric.
%

In figure \ref{fig:util_en} the utility gained for each budget is shown. It can be seen that the GA method performs at least equally good as the $5\%$ exact method. That happened only in the case where the robots had $25\%$ of the maximum energy. In all the other cases the GA performed better. Comparing the two different population generation methods, one can see that the more sophisticated approach gives statistically better results. The random method performs close to the $5\%$ optimal.

\begin{table}[]
\centering
\caption{Average time(s) for different algorithms and different budgets}
\label{tab:miqp_limited_energy_time}
\begin{tabular}{lllll}
\hline \hline
\multirow{2}{*}{Algorithm} & \multicolumn{4}{c}{Budget} \Tstrut \\
& 100\% & 75\%  & 50\%  & 25\% \Bstrut \\ \hline
CTOP 5\% &695.60 &267.94 &3611.76 &3764.12 \Tstrut \\
GA-RANDOM & \textbf{0.721606} & \textbf{0.625391} & \textbf{0.362355} & \textbf{0.141678} \\
St.Dev. & 0.0742465 & 0.0607859 & 0.0346406 & 0.0147053 \\
GA-NNRASP & 0.837791 & 0.763285 & 0.389283 & 0.165087 \\
St.Dev. & 0.120398 & 0.102063 & 0.0483728 & 0.0244343 \Bstrut \\
\hline
\end{tabular}
\end{table}

Table \ref{tab:miqp_limited_energy_time} the timing results of the limited budget experiments are presented. As with all the previous cases the GA-Random method performed better. In the worst case it was more than 300 times faster than the $5\%$ MIQP solution. The GA with NNRASP population generation method performed constantly a bit slower than the Random generation method.

\section{Conclusions}
\label{sec:conclusion}
This paper studies the online task scheduling problem for sensing missions performed by multiple vehicles. Such problems can be solved using the correlated team orienteering problem. The CTOP is designed for such problems as it takes into account the correlation of sensed information between various sensing points, while keeping the resource usage constrained. This work presents a heuristic for solving the CTOP that is based on a genetic algorithm. Initially, it uses a method from the literature to tune the GA parameters. The heuristic is then compared against a mixed integer quadratic programming solution. It is shown that in terms of utility the heuristic performs usually better than a near optimal MIQP solution. In terms of solution time complexity the heuristic performs at least 300 times better than the MIQP. The average execution time for the studied instances was in the worst case less than a second. These results make the proposed GA heuristic suitable for online usage. Future work would involve testing the heuristic in larger, more complex instances where the exact solution is inapplicable. Finally, it would be worth exploring the coordination capabilities of a heterogeneous team with different capabilities and objectives using a solution produced by a single instance of the proposed heuristic.

\bibliographystyle{IEEEtran}
\bibliography{paper} 

\end{document}